\documentclass[letterpaper, 10 pt, conference]{ieeeconf}  %

\IEEEoverridecommandlockouts                              %

\usepackage{graphicx} %
\usepackage{amsmath} %
\usepackage{amssymb}  %
\usepackage{amsfonts}  %
\usepackage{hyperref}
\usepackage{ifthen}
\usepackage{booktabs}\usepackage{multirow}
\usepackage{verbatim}
\usepackage{xcolor}
\usepackage{bm}
\usepackage[percent]{overpic}

\newboolean{useJPEG}
\setboolean{useJPEG}{true}  %
\usepackage{esvect}

\newcommand{\etal}{\emph{et al.}}
\newcommand{\eg}{\emph{e.g.}}

\newcommand{\viz}{\emph{viz.}}

\newcommand{\R}{\mathbb{R}}
\newcommand{\bfp}{\ensuremath{{\bf p}}}
\newcommand{\bfk}{\ensuremath{{\bf k}}}

\title{\LARGE \bf
Camera-to-Robot Pose Estimation from a Single Image
}

\author{\large\textbf{Timothy E.~Lee}$^{1,2}$, \textbf{Jonathan Tremblay}$^{1}$,  
\textbf{Thang To}$^{1}$, \textbf{Jia Cheng}$^{1}$, \\ \textbf{Terry Mosier}$^{1}$, \textbf{Oliver Kroemer}$^{2}$, \textbf{Dieter Fox}$^{1}$,
and \textbf{Stan Birchfield}$^{1}$,\\
$^{1}$NVIDIA: {\small\texttt{\{jtremblay, thangt, jicheng, tmosier, dieterf, sbirchfield\}@nvidia.com}}\\
$^{2}$Carnegie Mellon University, The Robotics Institute: {\small\texttt{\{timothyelee, okroemer\}@cmu.edu}}\\
\thanks{Work was completed while the first author was an intern at NVIDIA.}%
}

\begin{document}

\maketitle
\thispagestyle{empty}
\pagestyle{empty}

\begin{abstract}
We present an approach for estimating the pose of an external camera with respect to a robot using a single RGB image of the robot.
The image is processed by a deep neural network to detect 2D projections of keypoints (such as joints) associated with the robot.
The network is trained entirely on simulated data using domain randomization to bridge the reality gap.
Perspective-$n$-point (P$n$P) is then used to recover the camera extrinsics, assuming that the camera intrinsics and joint configuration of the robot manipulator are known.  
Unlike classic hand-eye calibration systems, our method does not require an off-line calibration step.
Rather, it is capable of computing the camera extrinsics from a single frame, thus opening the possibility of on-line calibration.
We show experimental results for three different robots and camera sensors, demonstrating that our approach is able to achieve accuracy with a single frame that is comparable to that of classic off-line hand-eye calibration using multiple frames.  
With additional frames from a static pose, accuracy improves even further.
Code, datasets, and pretrained models for three widely-used robot manipulators are made available.\footnote{\scriptsize \url{https://research.nvidia.com/publication/2020-03_DREAM}}
\end{abstract}

\section{INTRODUCTION}\label{sec:intro}

Determining the pose of an externally mounted camera is a fundamental problem for robot manipulation,
because it is necessary to transform measurements made in camera space to the robot's task space.
For robots operating in unstructured, dynamic environments---performing tasks such as object grasping and manipulation, human-robot interaction, and collision detection and avoidance---such a transformation allows visual observations to be readily used for control.

The classic approach to calibrating an externally mounted camera is to fix a fiducial marker (\eg, ArUco~\cite{garrido2014pr:aruco}, ARTag~\cite{fiala2005cvpr:artag}, or AprilTag~\cite{olson2011icra:apriltags}) to the end effector, collect several images, then solve a homogeneous linear system for the unknown transformation~\cite{fassi2005jrs:axxb}.
This approach is still widely used due to its generality, flexibility, and the availability of open-source implementations in the Robot Operating System (ROS).
However, such an approach requires the somewhat cumbersome procedure of physically modifying the end effector, moving the robot to multiple joint configurations to collect a set of images (assuming the marker-to-wrist transformation is not known), running an off-line calibration procedure, and (optionally) removing the fiducial marker.  Such an approach is not amenable to online calibration, because if the camera moves with respect to the robot, the entire calibration procedure must be repeated from scratch.

A more recent alternative is to avoid directly computing the camera-to-robot transform altogether, and instead to rely on an implicit mapping that is learned for the task at hand.  For example, Tobin \etal~\cite{tobin2017iros:dr} use deep learning to map RGB images to world coordinates on a table, assuming that the table at test time has the same dimensions as the one used during training.
Similarly, Levine \etal~\cite{levine2018learning} learn hand-eye coordination for grasping a door handle, using a large-scale setup of real robots for collecting training data.
In these approaches the learned mapping is implicit and specific to the task/environment, thus preventing the mapping from being applied to new tasks or environments without retraining.

We believe there is a need for a general-purpose tool that performs online camera-to-robot calibration without markers.  
With such a tool, a researcher could set up a camera (\eg, on a tripod), and then immediately use object detections or measurements from image space for real-world robot control in a task-independent manner, without a separate offline calibration step.
Moreover, if the camera subsequently moved for some reason (\eg, the tripod were bumped accidentally), there would be no need to redo the calibration step, because the online calibration process would automatically handle such disturbances.

In this paper, we take a step in this direction by presenting a system for solving camera pose estimation from a single RGB image.
We refer to our framework as \emph{DREAM} (for Deep Robot-to-camera Extrinsics for Articulated Manipulators). 
We train a robot-specific deep neural network to estimate the 2D projections of prespecified keypoints (such as the robot's joints) in the image.
Combined with camera intrinsics, the robot joint configuration, and forward kinematics, the camera extrinsics are then estimated using Perspective-$n$-Point (P$n$P)~\cite{lepetit2009ijcv:epnp}.
The network is trained entirely on synthetic images, relying on domain randomization~\cite{tobin2017iros:dr} to bridge the reality gap.
To generate these images, we augmented our open-source tool, NDDS \cite{to2018ndds}, to allow for scripting robotic joint controls and to export
metadata about specific 3D locations on a 3D mesh. 

This paper makes the following contributions:
\begin{itemize}
\item We demonstrate the feasibility of computing the camera-to-robot transformation from a single image of the robot, \emph{without fiducial markers}, using a deep neural network trained only on synthetic data.  
\item We show that the resulting accuracy with a single real image frame is comparable to that of classic hand-eye calibration using multiple frames.  
For a static camera, accuracy further improves with multiple image frames.
\item Quantitative and qualitative results are shown for three different robot manipulators (Franka Emika's Panda, Kuka's LBR iiwa 7 R800, and Rethink Robotics' Baxter) and a variety of cameras.
\end{itemize}
The simplicity of our approach enables real-time pose estimation on a desktop computer with an NVIDIA Titan Xp GPU, without manual optimization.

\section{APPROACH}\label{sec:method}

\begin{figure*}
    \centering
     \includegraphics[width=\linewidth]{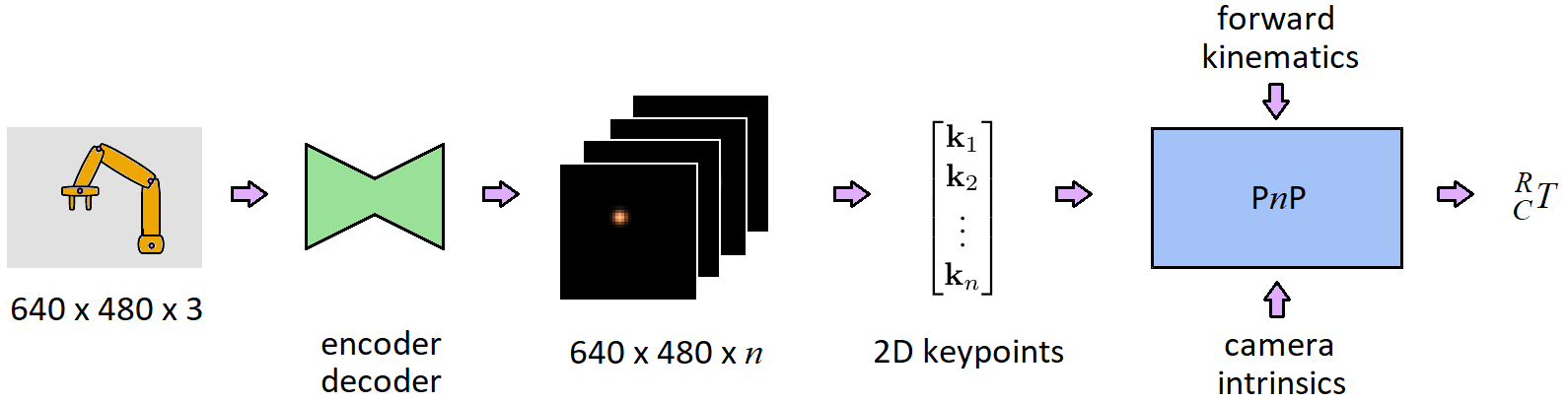}		
    \caption{The DREAM framework. 
    A deep encoder-decoder neural network takes as input an RGB image of the robot from an externally-mounted camera, and it outputs $n$ belief maps (one per keypoint). 
    The 2D peak of each belief map is then extracted and used by P$n$P, along with the forward kinematics and camera intrinsics, to estimate the camera-to-robot pose, ${_C^R}T$.} 
    \label{fig:framework_pipeline}
\end{figure*}

An externally mounted camera observes $n$ keypoints $\bfp_i \in \R^3$ on various robot links.  These keypoints project onto the image as $\bfk_i \in \R^2$, $i=1,\ldots,n$.  Some of these projections may be inside the camera frustum, whereas others may be outside.  We consider the latter to be invisible/inaccessible, whereas the former are visible, regardless of occlusion.\footnote{The network learns to estimate the positions of occluded keypoints from the surrounding context; technically, since the keypoints are the robot joints (which are inside the robot links), they are always occluded.}
The intrinsic parameters of the camera are assumed known.

Our proposed two-stage process for solving the problem of camera-to-robot pose estimation from a single RGB image frame is illustrated in Fig.~\ref{fig:framework_pipeline}.
First, an encoder-decoder neural network processes the image to produce a set of $n$ belief maps, one per keypoint.
Then, Perspective-$n$-Point (P$n$P)~\cite{lepetit2009ijcv:epnp} uses the peaks of these 2D belief maps, along with the forward kinematics of the robot and the camera intrinsics, to compute the camera-to-robot pose, ${^R_C}T$.
Note that the network training depends only on the images, not the camera; therefore, after training, the system can be applied to any camera with known intrinsics.
We restrict $n \ge 4$ for stable P$n$P results.

\subsection{Network Architecture}

Inspired by recent work on object pose estimation~\cite{zakharov2019dpod,peng2019pvnet,hu2019segmentation}, we use an auto-encoder network to detect the keypoints.  
The neural network takes as input an RGB image of size $w \times h \times 3$, and it outputs an $\alpha w \times \alpha h \times n$ tensor, where $w = 640$, $h=480$, and $\alpha \in \{1, \frac{1}{2}, \frac{1}{4}\}$, depending on the output resolution used. 
This output captures a 2D belief map for each keypoint, where pixel values represent the likelihood that the keypoint is projected onto that pixel.

The encoder consists of the convolutional layers of VGG-19 \cite{simonyan2015iclr:vgg} pretrained on ImageNet. 
The decoder (upsampling) component is composed of four 2D transpose convolutional layers (stride = 2, padding = 1, output padding = 1), and  each layer is followed by a normal $3 \times 3$ convolutional layer and ReLU activation layer.
We also experimented with a ResNet-based encoder, \viz, our reimplementation of~\cite{xiao2018simple}, with the same batch normalization, upsampling layers, and so forth as described in the paper.

The output head is composed of 3 convolutional layers ($3\times3$, stride = 1, padding = 1) with ReLU activations with 64, 32, and $n$ channels, respectively. 
There is no activation layer after the final convolutional layer.
The network is trained using an L2 loss function comparing the output belief maps with ground truth belief maps, where the latter are generated using $\sigma=2$~pixels to smooth the peaks.

\subsection{Pose Estimation}

Given the 2D keypoint coordinates, robot joint configuration with forward kinematics, and camera intrinsics, P$n$P~\cite{lepetit2009ijcv:epnp} is used to retrieve the pose of the robot, 
similar to \cite{tremblay2018pose,tekin2018cvpr:objpose,zakharov2019dpod,peng2019pvnet,hu2019segmentation}. 
The keypoint coordinates are calculated as a weighted average of values near thresholded peaks in the output belief maps (threshold = 0.03), after first applying Gaussian smoothing to the belief maps to reduce the effects of noise.
The weighted average allows for subpixel precision.

\subsection{Data Generation}

The network is trained using only synthetic data with domain randomization (DR) and image augmentation~\cite{buslaev2018arx:albumentations}.
Despite the traditional challenges with bridging the reality gap, we find that our network generalizes well to real-world images, as we will show in the experimental results.
To generate the data we used our open-source NDDS tool~\cite{to2018ndds}, which is 
a plugin for the UE4 game engine. 
We augmented NDDS to export 3D/2D keypoint locations and robot joint angles, as well as to allow control of the robot joints. 

The synthetic robot was placed in a simple virtual 3D scene in UE4, viewed by a virtual camera.
Various randomizations were applied:
1) The robot's joint angles were randomized within the joint limits.
2) The camera was positioned freely in a somewhat truncated  hemispherical shell around the robot, with azimuth ranging from $-135^\circ$~to $+135^\circ$~(excluding the back of the robot), 
elevation from $-10^\circ$ to $75^\circ$, and distance from 75~cm to 120~cm; the optical axis was also randomized within a small cone. 
3) Three scene lights were positioned and oriented freely while randomizing both intensity and color.
4) The scene background was randomly selected from the COCO dataset~\cite{Lin2014COCO}. 
5) 3D objects from the YCB dataset~\cite{calli2015icar:ycb} were randomly placed, similar to flying distractors~\cite{tremblay2018wad:car}.
6) Random color was applied to the robot mesh.

\begin{figure}
    \centering
    \begin{tabular}{cc}
        \includegraphics[width=0.45\linewidth]{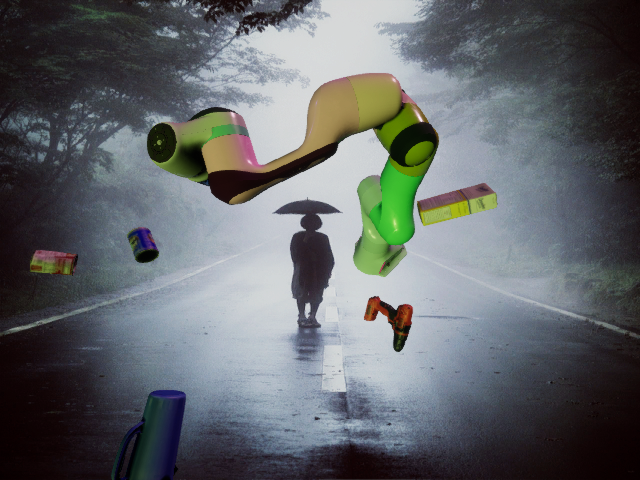} &
        \includegraphics[width=0.45\linewidth]{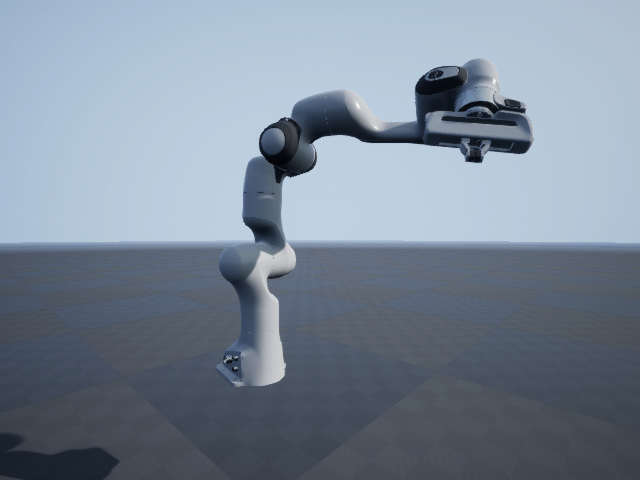} \\
        \includegraphics[width=0.45\linewidth]{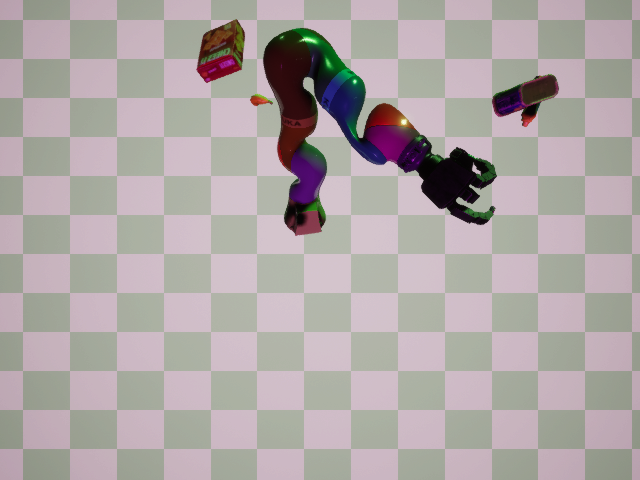} &
        \includegraphics[width=0.45\linewidth]{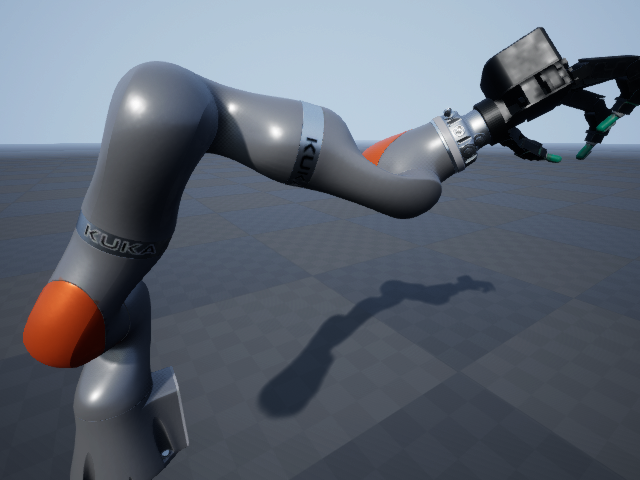} \\
        \includegraphics[width=0.45\linewidth]{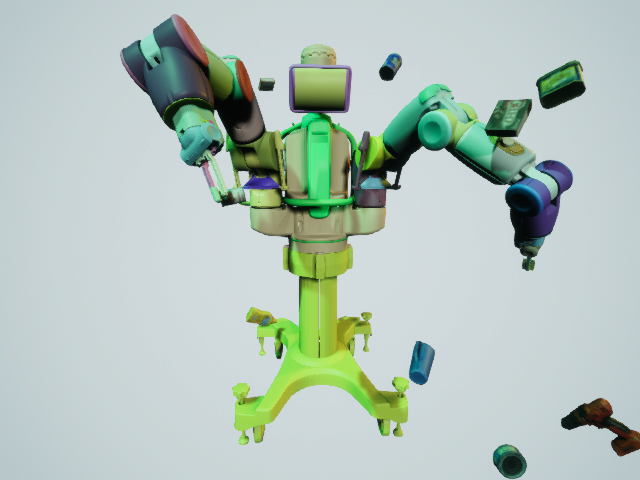} &
        \includegraphics[width=0.45\linewidth]{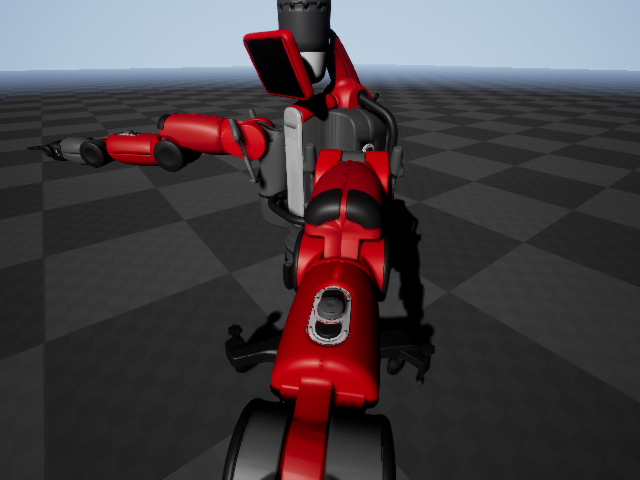} \\
				domain-randomized (DR) & non-DR
    \end{tabular}
    \caption{Synthetic training images for the three robot models:  Franka Panda (top), Kuka LBR with Allegro hand (middle), and Rethink Baxter (bottom).}
    \label{fig:synthetic_dr_ndr_images}
\end{figure}

Figure~\ref{fig:synthetic_dr_ndr_images} shows representative images from synthetic datasets generated by our approach.
Domain randomized (DR) datasets were used for training and testing; datasets without domain randomization (non-DR) were used for testing to assess sim-to-sim generalization.
The DR training data was generated using the publicly available robot models, whereas the non-DR test data used models that were artistically improved to be more photorealistic.

\section{EXPERIMENTAL RESULTS}\label{sec:results}

In this section we evaluate our DREAM method both in simulation and in the real world.  
We seek to answer the following questions:  1) How well does the synthetic-only training paradigm transfer to real-world data?  2) What is the impact of various network architectures?  3) What accuracy can be achieved with our system, and how does it compare to traditional hand-eye calibration?

\subsection{Datasets}

We collected several real-world datasets in our lab using various RGBD sensors.
Since our DREAM algorithm processes only RGB images, the depth images were captured solely to facilitate ground truth camera poses via DART~\cite{schmidt2014dart}, a depth-based articulated tracking algorithm.
DART was initialized manually and monitored in real-time during data collection to ensure tracking remained stable and correct, by viewing the projection of the robot coordinate frames onto the camera's RGB images via the ROS visualization tool (RViz).

\textbf{Panda-3Cam.}  \indent %
For this dataset, the camera was placed on a tripod aimed at the Franka Emika Panda robot arm.
The robot was moved to five different joint configurations, at which the camera collected data for approximately five seconds each, followed by manual teleoperation to provide representative end effector trajectories along cardinal axes, as well as along a representative reaching trajectory.
During data collection, neither the robot base nor the camera moved. 
The entire capture was repeated using \emph{three different cameras} utilizing different modalities: Microsoft XBox~360 Kinect (structured light), Intel RealSense D415 (active stereo), and Microsoft Azure Kinect (time-of-flight). 
All images were recorded natively at $640 \times 480$ resolution at 30~fps, except for the Azure Kinect, which was collected at $1280 \times 720$ and downsampled to $640 \times 480$ via bicubic interpolation.
This dataset consists of 17k total image frames divided approximately equally between the three cameras. 

\textbf{Panda-Orb.} \indent To evaluate the method's ability to handle a variety of camera poses, additional data was captured of the Panda robot.  
The RealSense camera was again placed on a tripod, but this time at 27 different positions in a roughly \emph{orbital} motion around the robot (namely, 9 different azimuths, ranging from roughly -180$^\circ$ to +180$^\circ$, and for each azimuth two different elevations approximately 30$^\circ$ and 45$^\circ$, along with a slightly closer depth at 30$^\circ$).   
For each camera pose, the robot was commanded using Riemannian Motion Policies~(RMPs)~\cite{ratliff2018arx:rmp,cheng2018wafr:rmpflow} to perform the same joint motion sequence of navigating between 
10 waypoints defined in both task and joint space. 
The dataset consists of approximately 40k image frames.

\subsection{Metrics}

Metrics were computed on both 2D and 3D.
For the 2D evaluation, the percentage of correct keypoints (PCK)~\cite{tremblay2017icra:cube} below a certain threshold was calculated, as the threshold varied.
All keypoints whose ground truth was within the camera's frustum were considered, regardless of whether they were occluded.
For 3D evaluation of the accuracy of the final camera-to-robot pose, the average distance (ADD)~\cite{hinterstoisser2012accv:linemod,xiang2018rss:posecnn} was calculated, which is the average Euclidean distance of all 3D keypoints to their transformed versions, using the estimated camera pose as the transform.
ADD is a principled way, based upon Euclidean geometry, to combine rotation and translation errors without introducing an arbitrary weighting between them.
As with PCK, the percentage of keypoints with ADD lower than a threshold was calculated, as the threshold varied.
For ADD, all keypoints were considered.
In both cases, averages were computed over keypoints over all image frames.

\subsection{Training and Simulation Experiments}

For comparison, we trained three versions of our DREAM network.  As described earlier, the architecture uses either a VGG- or ResNet-based encoder, and the decoder outputs either full (F), half (H), or quarter (Q) resolution.
Each neural network was trained for 50~epochs using Adam~\cite{kingma2015iclr:adam} with 1.5e-4 learning rate and 0.9 momentum. 
Training for each robot used approximately 100k synthetic DR images.
The best-performing weights were selected according to a synthetic validation set.

\begin{figure}
    \centering
    \begin{tabular}{cc}
        \hspace{-1em}        \includegraphics[width=0.5\linewidth]{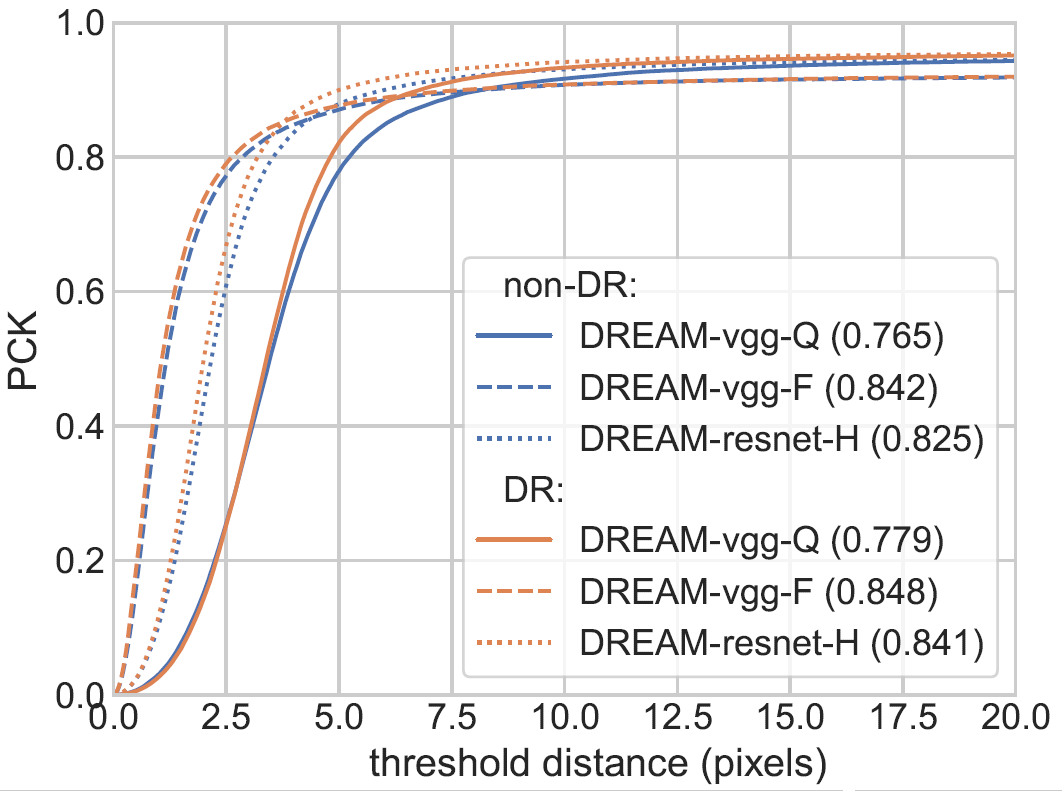} &
        \hspace{-1em} \includegraphics[width=0.5\linewidth]{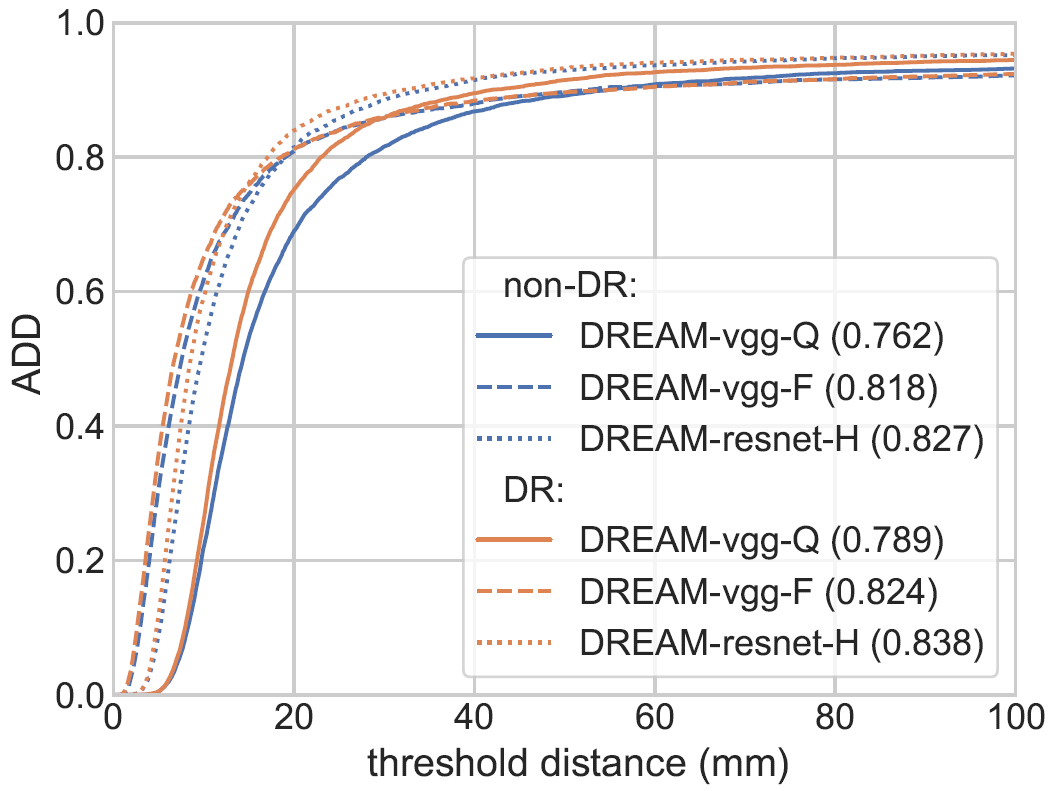}
    \end{tabular}
    \caption{PCK (left) and ADD (right) results for three different variants of our DREAM network on the two simulated datasets.  The numbers in parentheses are the area under the curve (AUC).}
    \label{fig:results_sim}
\end{figure}

As a baseline, Fig.~\ref{fig:results_sim}
compares these versions on two simulated datasets, one with domain-randomization (DR) and the other without (non-DR).  
The improvement due to increasing resolution is clear, but different architectures have only minimal impact for most scenarios.

\subsection{Real-world Experiments}

\begin{figure}
    \centering
    \includegraphics[width=\linewidth]{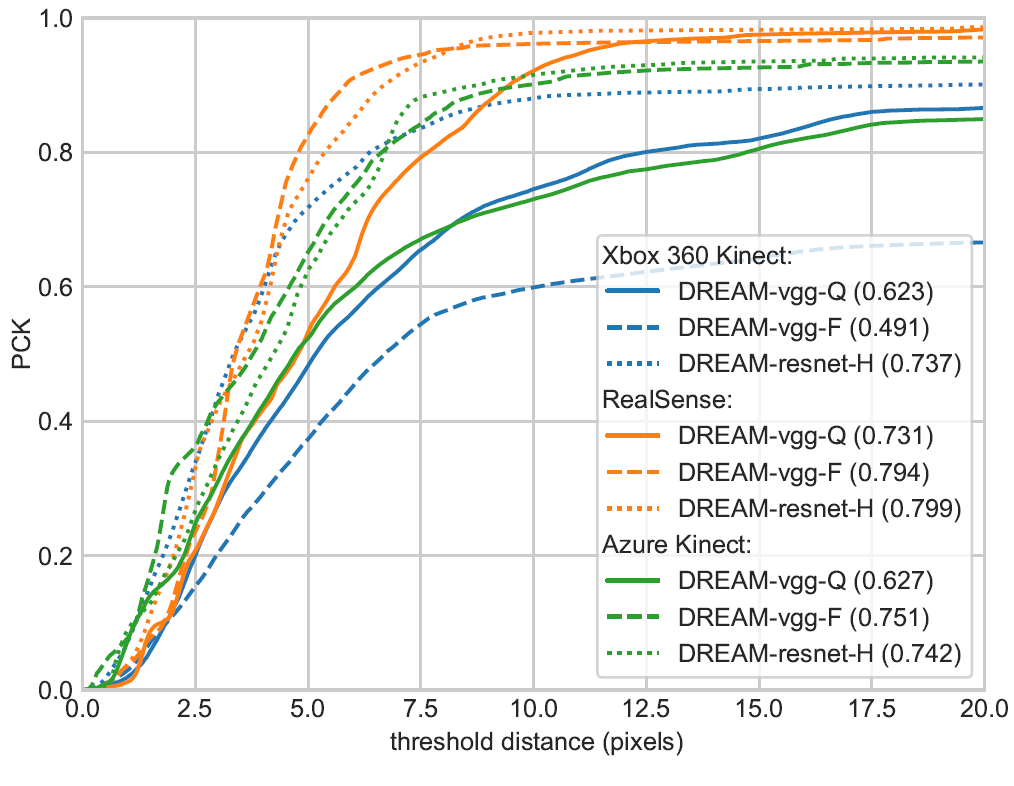} \\
    \includegraphics[width=\linewidth]{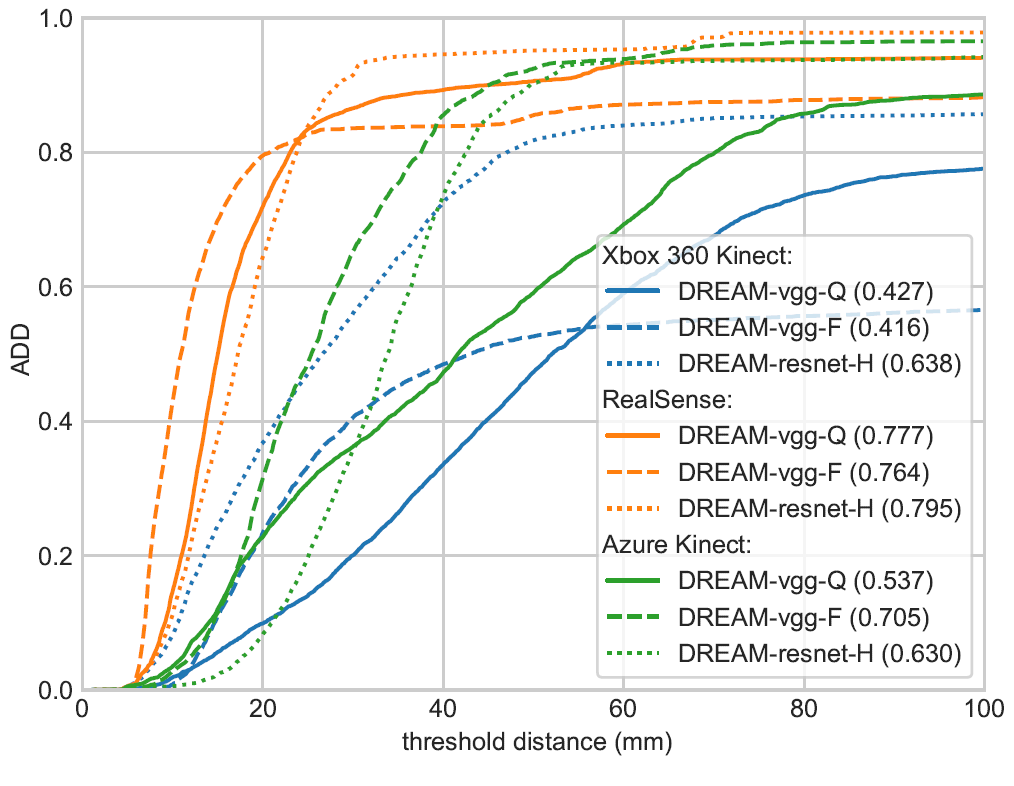}
    \caption{PCK (top) and ADD (bottom) results on the real Panda-3Cam dataset.}
    \label{fig:results_F3}
\end{figure}

Results comparing the same three networks on the Panda-3Cam dataset are shown in Fig.~\ref{fig:results_F3}.
Encouragingly, these results show that our training procedure is able to bridge the reality gap:  There is only a modest difference between the best performing network on simulated and real data.

For 3D, it is worth noting that the standard deviation of ground truth camera pose from DART was 1.6~mm, 2.2~mm, and 2.9~mm, respectively, for the XBox~360 Kinect (XK), RealSense (RS), and Azure Kinect (AK) cameras, respectively.
The degraded results for the Azure Kinect are due to DART's sensitivity to noise in the time-of-flight-based depth, rather than to DREAM itself.
On the other hand, the degraded results for XBox 360 Kinect are due to the poor RGB image quality from that sensor.

Ultimately, the goal is to be able to place the camera at an arbitrary pose aimed at a robot, and calibrate automatically.
To measure DREAM's ability to achieve this goal, we evaluated the system on the Panda-Orb dataset containing multiple camera poses.
These results, alongside those of the previous experiments, are shown in Tab.~\ref{tab:results_pandaorb_kuka}.
For this table we used DREAM-vgg-F, since the other architectures performed similarly as before.

\begin{table}
    \centering
    \caption{Results at different thresholds of the same network (DREAM-vgg-F) on various datasets (and camera sensors).  All but the last row are taken from Figs.~\ref{fig:results_sim} and~\ref{fig:results_F3}.}

    \begin{tabular}{c||c|c|c||c|c|c}
    & \multicolumn{3}{c||}{PCK @~(pix)} 
    & \multicolumn{3}{c}{ADD @~(mm)} 
    \\
    Dataset & 2.5 & 5.0 & 10.0 & 20 & 40 & 60 \\
    \hline
    Sim. DR &  0.79   &  0.88   &   0.90   &  0.81  &  0.88  &  0.90 \\
    Sim. non-DR &  0.77   &  0.87   &  0.90    &  0.80  &  0.88  & 0.90\\
    Panda-3Cam (XK) & 0.15  &  0.37   &  0.59    &  0.23  & 0.48   &0.54\\
    Panda-3Cam (RS) & 0.24 &0.83 &0.96 & 0.80& 0.83&0.87\\
    Panda-3Cam (AK) & 0.36 &0.65 &0.90 & 0.32 &0.86 &0.94\\
    Panda-Orb (RS) & 0.28 & 0.67 & 0.83 & 0.57 & 0.77 & 0.80 \\

    \end{tabular}
    \label{tab:results_pandaorb_kuka}
\end{table}

\begin{figure*}
    \centering
    \begin{tabular}{ccc}
    \includegraphics[width=0.32\linewidth]{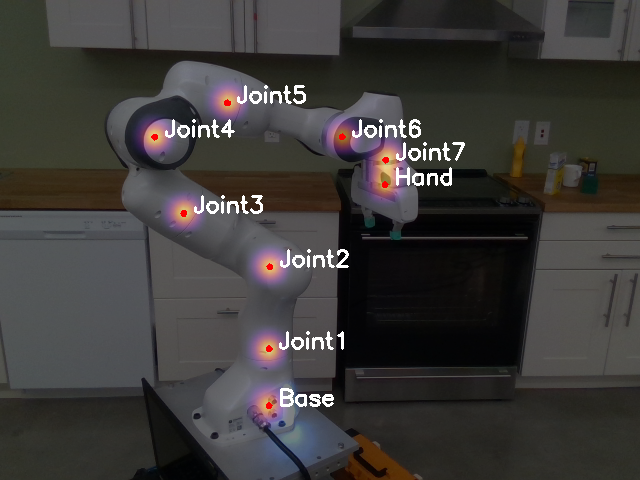} &
    \includegraphics[width=0.32\linewidth]{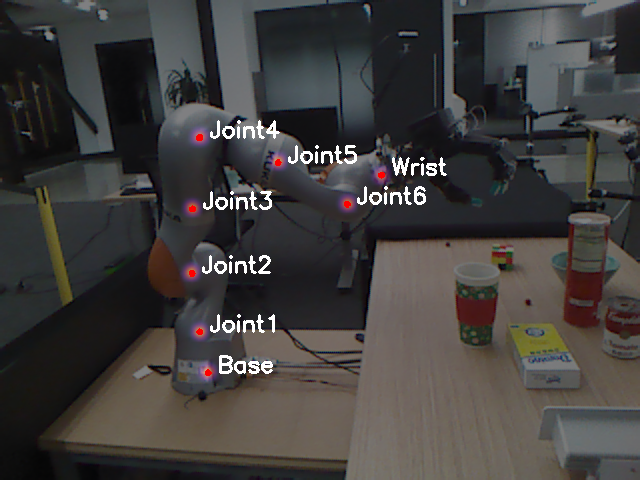} &
    \includegraphics[width=0.32\linewidth]{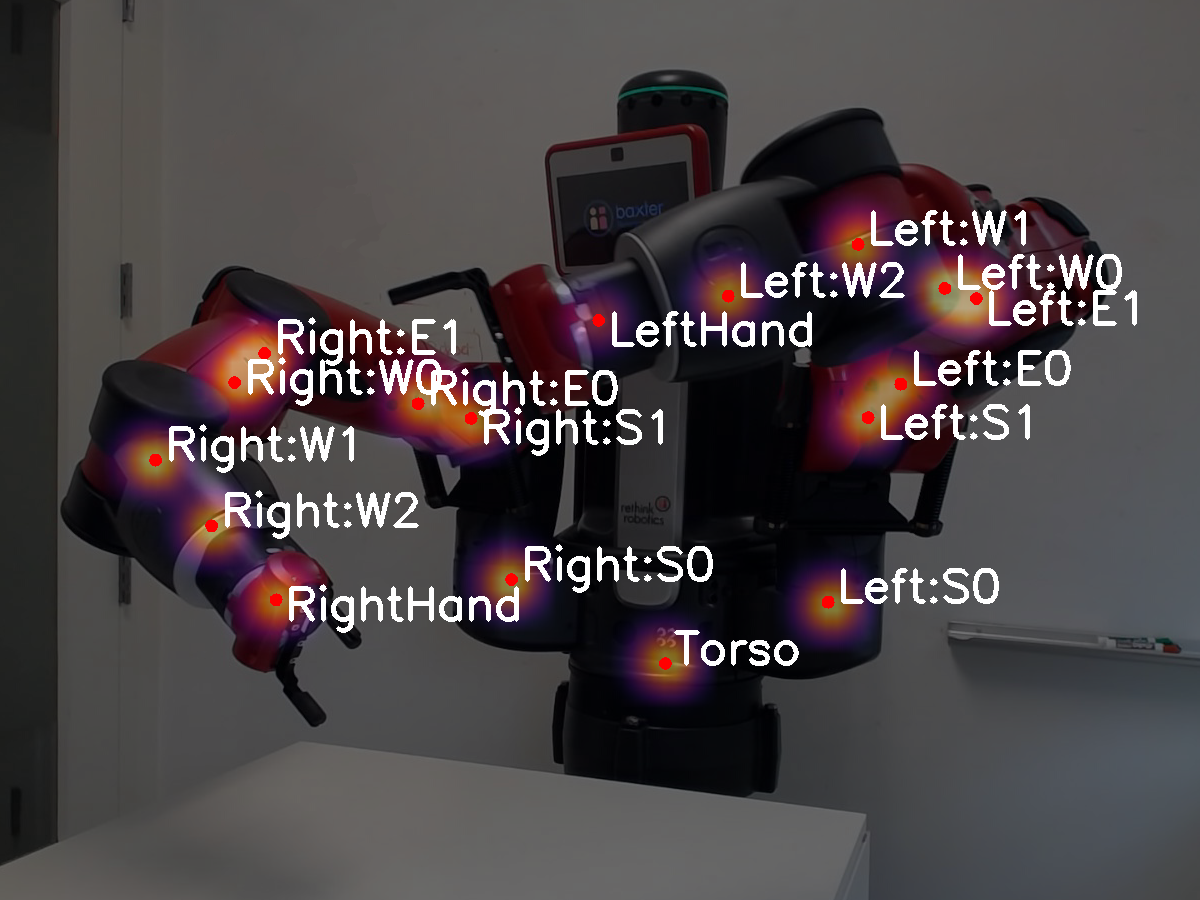}  %
    \end{tabular}
    \caption{Keypoint belief maps (red dots indicate peaks) detected by DREAM in RGB images of three different robots (taken by three different cameras).  From left to right:  Franka Emika Panda (Intel RealSense D415), Kuka LBR iiwa (Logitech C920 webcam), and Rethink Baxter (cell phone camera).}
    \label{fig:threerobotkps}
\end{figure*}

We also trained DREAM on the Kuka LBR iiwa and Baxter robots.  
While the former is similar to the Panda, the latter is more difficult due to symmetry that causes the two arms to be easily confused with one another.  To alleviate this problem, we had to restrict the azimuth range from -45$^\circ$ to +45$^\circ$.  
Although we did not perform quantitative analysis on either robot, qualitatively we found that the approach works about the same for these robots as it does for the Panda.
The detected keypoints overlaid on images of the three robots, using three different RGB cameras, are shown in Fig.~\ref{fig:threerobotkps}.
Although in principle the keypoints could be placed anywhere on the robot, we simply assign keypoints to the joints according to each robot's URDF (Unified Robot Description Format).

\subsection{Comparison with Hand-Eye Calibration}

The goal of our next experiment was to assess the accuracy of DREAM versus traditional hand-eye calibration (HEC).
For the latter, we used the \texttt{easy\_handeye} ROS package\footnote{\url{https://github.com/IFL-CAMP/easy_handeye}} to track an ArUco fiducial marker~\cite{garrido2014pr:aruco} attached to the Panda robot hand.

The XBox 360 Kinect was mounted on a tripod, and the robot arm was moved to a sequence of $M=18$ different joint configurations, stopping at each configuration for one second to collect data.
Neither the camera nor the base of the robot moved.
The fiducial marker was then removed from the hand, and the robot arm was moved to a different sequence of $M$ joint configurations.  
The joint configurations were selected favorably to ensure that the fiducial markers and keypoints, respectively, were detected in the two sets of images.
As before, DART with manual initialization was used for ground truth.

Although our DREAM approach works with just a single RGB image, it can potentially achieve greater accuracy with multiple images by simply feeding all detected keypoints (from multiple frames) to a single P$n$P call.
Thus, to facilitate a direct comparison with HEC, we applied DREAM to $m \ge 1$ images from the set of $M$ images that were collected.
Similarly, we applied HEC to $m \ge 3$ images from the set.
Both algorithms were then evaluated by comparing the estimated pose with the ground truth pose via ADD.
For both algorithms, we selected $\binom{M}{m}$ possible combinations when evaluating the algorithm on $m$ images, to allow the mean, median, min, and max to be computed.
To avoid unnecessary combinatorial explosion, whenever this number exceeded $N=2500$, we randomly selected $N$ combinations rather than exhaustive selection. 

\begin{figure}
    \centering
    \hspace*{-2ex}\includegraphics[width=1.1\linewidth]{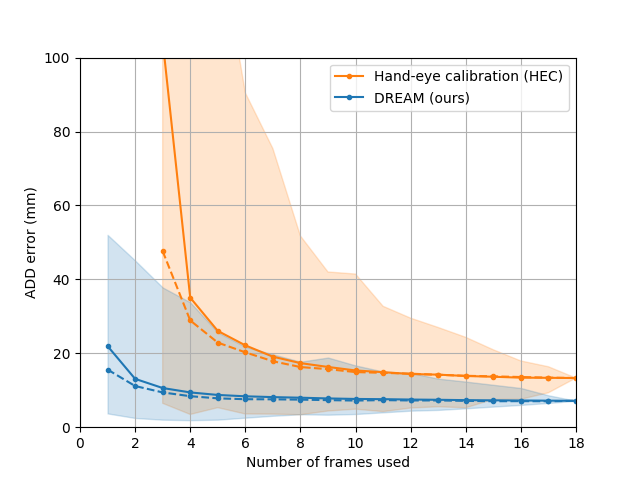}
    \caption{DREAM vs.\ HEC, measured by ADD as a function of the number of image frames used for calibration. Shown are the mean (solid line), median (dashed line), and min/max (shaded area), computed over different image combinations.  DREAM requires only a single image frame but achieves greater accuracy with more images.}
    \label{fig:dream_vs_hec}
\end{figure}

Results of this head-to-head comparison are shown in Fig.~\ref{fig:dream_vs_hec}.
Note that HEC is unable to estimate the camera pose when $m < 3$, whereas DREAM works with just a single image.
As the number of images increases, the estimated pose from both DREAM and HEC improves, depending somewhat on the robot configurations used.
In all cases, DREAM performs as well or better than HEC.
(Note, however, that HEC results would likely improve from manually, rather than randomly, selecting image frames.)

\subsection{Measuring Workspace Accuracy}

In the final experiment we evaluated the accuracy of DREAM's output with respect to the workspace of the robot.
The RealSense camera was placed on a tripod facing the Panda robot reaching toward an adjustable-height table on which were placed five fiducial markers.
A head-to-head comparison of the camera poses computed by DART, DREAM, and HEC was conducted by transforming each fiducial marker's pose from the camera's frame to the robot's frame by applying each algorithm's camera-to-robot pose estimate.
The robot was then commanded to move the end effector to a target position defined 10~cm directly above the marker, to avoid potential collision.
This process was repeated for ten target positions (5 markers, 2 table heights).
The Euclidean distance between the end effector's position in 3D was measured for each algorithm.
Note that in this experiment DART was not considered to be ground truth, but rather was compared against the other methods.

Results are shown in Tab.~\ref{tab:dartdreamhec}.
Even though DREAM is RGB-only, it performs favorably not only to HEC but also to the depth-based DART algorithm.
This is partly explained by the fact that the extrinsic calibration between the depth and RGB cameras is not perfect.
Note that DREAM's error is similar to that of our previous work~\cite{tremblay2018pose} upon which it is based, where we showed that an error of approximately 2~cm for object pose estimation from RGB is possible.

\begin{table}
    \centering
    \caption{Euclidean error between the robot's actual reached position and the commanded position, using the camera pose estimated by each of the three methods.}
    \begin{tabular}{c||c|c|c}
    & DART & HEC & DREAM (ours) \\
    \hline
    camera & depth & RGB & RGB \\
    no. frames & 1 & 10 & 1 \\
    \hline
    min error (mm) & 10.1 & \textbf{9.4} & 20.2 \\
    max error (mm) & 44.3 & 51.3 & \textbf{34.7} \\ 
    mean error (mm) & \textbf{21.4} & 28.2 & 27.4 \\
    std error (mm) & 12.3 & 14.2 & \textbf{4.7} \\
    \end{tabular}

    \label{tab:dartdreamhec}
\end{table}

\section{PREVIOUS WORK}\label{sec:lit_review}

Relationship to previous work is considered in this section.

\textbf{Object pose estimation.}\,\, 
In robotics applications, it is not uncommon for objects to be detected via fiducial markers~\cite{liu2018:robotsafe,kim2019icra:insertion,tian2019icra:fog}.
Even so, there is growing interest in the problem of markerless object pose estimation in both the robotics and computer vision 
communities \cite{tremblay2018pose,hinterstoisser2012accv:linemod,hodan2017wacv:tless,zakharov2019dpod,xiang2018rss:posecnn,hu2019segmentation,peng2019pvnet,Sundermeyer_2018_ECCV,tekin2018cvpr:objpose}, building upon work in keypoint detection for human pose estimation \cite{wei2016cvpr:cpm,cao2017cvpr:mppaf,xiao2018simple,li2019rethinking,sun2019deep}. 
Recent leading methods are similar to the approach proposed here:  A network is trained to predict object keypoints in the 2D image, followed by  
P$n$P \cite{lepetit2009ijcv:epnp} to estimate the pose of the object in the camera coordinate frame \cite{tremblay2018pose,peng2019pvnet,tan2017arx:objpose,hu2019segmentation,tekin2018cvpr:objpose,dhall2019arx:cones}, or alternatively, a deformable shape model is fit to the detect keypoints~\cite{pavlakos2017icra:pose}.
Indeed, our approach is inspired by these methods.
Our approach builds upon the findings of Peng \etal~\cite{peng2019pvnet}, who showed that regressing to keypoints on the object is better than regressing to vertices of an enveloping cuboid.
Other methods have regressed directly to the pose \cite{xiang2018rss:posecnn,Sundermeyer_2018_ECCV}, but care must be taken not to bake the camera intrinsics into the learned weights.

\textbf{Robotic camera extrinsics.}\,\, 
Closely related to the problem of estimating the camera-to-object pose (just described) is that of estimating the camera-to-robot pose.
The classic solution to this problem is known as hand-eye calibration, in which a fiducial marker (such as ArUco~\cite{garrido2014pr:aruco}, ARTag~\cite{fiala2005cvpr:artag}, AprilTag~\cite{olson2011icra:apriltags}, or otherwise known object) is attached to the end effector and tracked through multiple frames.
Using forward kinematics and multiple recorded frames, the algorithm solves a linear system to compute the camera-to-robot transform 
\cite{park_robot_1994,ilonen2011icar,yang1994bmvc:calibrobcam}.
Similarly, an online calibration method is presented by Pauwels and Kragic~\cite{pauwels2016icra:onlinecalib}, in which the 3D position of a known object is tracked from nonlinear optimization over multiple frames.

An alternate approach is to move a small object on a table, command the robot to point to each location in succession, then use forward kinematics to calibrate~\cite{park2018arx:grasping}.
However, the accuracy of such an approach degrades as the robot moves away from the table used for calibration.
Aalerund \etal~\cite{aalerund2019sensors:calib} present a method for calibrating an RGBD network of cameras with respect to each other for robotics applications, but the camera-to-robot transforms are not estimated.

For completeness, we mention that, although our paper addresses the case of an externally mounted camera, another popular configuration is to mount the camera on the wrist~\cite{morrison2018rss:closeloop}, for which the classic hand-eye calibration approach applies~\cite{pauwels2016icra:onlinecalib}.
Yet another configuration is to mount the camera on the ceiling pointing downward, for which simple 2D correspondences are sufficient~\cite{feniello2014iros:progsyn,mahler2017rss:dexnet2,park2018arx:grasping}.

\textbf{Robotic pose estimation.}\,\, 
Bohg \etal~\cite{bohg2014robot} explore the problem of markerless robot arm pose estimation.
In this approach, a random decision forest is applied to a depth image to segment the links of the robot, from which the robot joints are estimated.
In follow-up work, Widmaier~\etal~\cite{widmaier2016robot} address the same problem but obviate the need for segmentation by directly regressing to the robot joint angles.
Neither of these approaches estimate the camera-to-robot transform.

The most similar approach to ours is the simultaneous work of Lambrecht and K{\"a}stner~\cite{lambrecht2019icar:mlpose,lambrecht2019rita:fspose}, in which a deep network is also trained to detect projected keypoints, from which the camera-to-robot pose is computed via P$n$P.  
A key difference is that our network is trained only on synthetic data, whereas theirs requires real and synthetic data.
In other recent work, Zuo \etal~\cite{zuo2019craves} also present a keypoint-based detection network.
But rather than use P$n$P, nonlinear optimization directly regresses to the camera pose and the unknown joint angles of a small, low-cost manipulator.
The network is trained using synthetic data, with domain adaptation to bridge the reality gap.  

\section{CONCLUSION}\label{sec:conclusion}

We have presented a deep neural network-based approach to compute the extrinsic camera-to-robot pose using a single RGB image.  
Compared to traditional hand-eye calibration, we showed that our DREAM method achieves comparable accuracy even though it does not use fiducial markers or multiple frames.
Nevertheless, with additional frames, our method is able to reduce the error even further.
We have presented quantitative results on a robot manipulator using images from three different cameras, and we have shown qualitative results on other robots using other cameras.
We believe the proposed method takes a significant step toward robust, online calibration.
Future work should be aimed at filtering results over time, computing uncertainty, and incorporating the camera pose into a closed-loop grasping or manipulation task.

\section*{ACKNOWLEDGMENTS}
We gratefully acknowledge Karl van Wyk, Clemens Eppner, Chris Paxton, Ankur Handa, and Erik Leitch for their help.
Many thanks also to Kevin Zhang and Mohit Sharma (Carnegie Mellon University).  

\bibliographystyle{IEEEtran}
\bibliography{refs}

\end{document}